%% file: main.tex
\documentclass[sigconf]{acmart}

\renewcommand\footnotetextcopyrightpermission[1]{} 

\usepackage{algorithmic}  
\usepackage{algorithm} 
\usepackage[algo2e]{algorithm2e} 
\usepackage{pifont,dsfont}
\usepackage{makecell}
\usepackage{amsfonts}
\usepackage{amsthm}
\usepackage{wrapfig,lipsum,booktabs}
\usepackage[export]{adjustbox}
\usepackage{xcolor,colortbl}

\usepackage{graphicx}
\usepackage{subfigure}
\usepackage{paralist}
\usepackage{refcount}
\usepackage{amsfonts}
\usepackage{caption}
\usepackage{makecell} 

\newtheorem{definition}{Definition}

\newcommand{\hide}[1]{}

\usepackage[most]{tcolorbox}
\definecolor{block-gray}{gray}{0.85}
\newtcolorbox{myquote}{colframe=black,boxrule=1pt,
colback=white,grow to right by=-1mm,grow to left by=-1mm,
boxsep=0pt,breakable}

\newtcolorbox{inside_myquote}{boxrule=0pt,
colback=block-gray,grow to right by=3mm,grow to left by=3mm,
top=0pt,bottom=0pt}

\AtBeginDocument{%
  \providecommand\BibTeX{{%
    \normalfont B\kern-0.5em{\scshape i\kern-0.25em b}\kern-0.8em\TeX}}}

\begin{document}


\title{Neural-Symbolic Reasoning over Knowledge Graphs:\\ A Survey from a Query Perspective}

\author{Lihui Liu}
\email{lihuil2@illinois.edu}
\affiliation{%
  \institution{Wayne State University}
  \city{Detroit}
  \state{Michigan}
  \country{USA}
}

\author{Zihao Wang}
\authornote{ZW contributed during his visit to UIUC in 2023-2024}
\email{zihaow@illinois.edu}
\authornotemark[0]
\affiliation{%
  \institution{University of Illinois at Urbana Champaign}
  \city{Urbana}
  \state{Illinois}
  \country{USA}
}

\author{Hanghang Tong}
\email{htong@illinois.edu}
\affiliation{%
  \institution{University of Illinois at Urbana Champaign }
  \city{Urbana}
  \state{Illinois}
  \country{USA}
}

\vspace{-2\baselineskip}

\renewcommand{\shortauthors}{Lihui Liu \& Hanghang Tong}

\begin{abstract}

\input{000abs.tex}

\vspace{-0.7\baselineskip}
\end{abstract}

\vspace{-1\baselineskip}
\vspace{-2\baselineskip}
\vspace{-2\baselineskip}


\maketitle


\section{Introduction}

\input{001intro.tex}

\section{Preliminary}\label{preliminary}

\input{002preliminary.tex}


\section{Reasoning for Single Hop Query}\label{single}

\input{003single}

\section{Reasoning for Complex Logic Query}\label{complex}
\input{005complex}

\section{Reasoning For Natural Language Query}\label{nlp}

\input{004language}

\section{LLM with Knowledge Graph Reasoning}\label{llm}
\input{006other}

\section{Conclusion and Future Directions}
\input{008conclusion}

\bibliographystyle{plain}
\bibliography{008reference.bib}


\end{document}

%% file: 000abs.tex
Knowledge graph reasoning is pivotal in various domains such as data mining, artificial intelligence, the Web, and social sciences. These knowledge graphs function as comprehensive repositories of human knowledge, facilitating the inference of new information. Traditional symbolic reasoning, despite its strengths, struggles with the challenges posed by incomplete and noisy data within these graphs. In contrast, the rise of Neural Symbolic AI marks a significant advancement, merging the robustness of deep learning with the precision of symbolic reasoning. This integration aims to develop AI systems that are not only highly interpretable and explainable but also versatile, effectively bridging the gap between symbolic and neural methodologies. Additionally, the advent of large language models (LLMs) has opened new frontiers in knowledge graph reasoning, enabling the extraction and synthesis of knowledge in unprecedented ways.
This survey offers a thorough review of knowledge graph reasoning, focusing on various query types and the classification of neural symbolic reasoning. Furthermore, it explores the innovative integration of knowledge graph reasoning with large language models, highlighting the potential for groundbreaking advancements. This comprehensive overview is designed to support researchers and practitioners across multiple fields, including data mining, AI, the Web, and social sciences, by providing a detailed understanding of the current landscape and future directions in knowledge graph reasoning.

%% file: 001intro.tex
A knowledge graph is a graph structure that contains a collection of facts, where nodes represent real-world entities, events, and objects, and edges denote the relationships between two nodes. It is a powerful tool for organizing and connecting information in a way that mimics human thought and learning. Since its debut in 2012,\footnote{\href{https://en.wikipedia.org/wiki/Knowledge_graph}{https://en.wikipedia.org/wiki/Knowledge\_graph}} a variety of knowledge graphs have been generated, including Freebase ~\cite{freebase}, Yago~\cite{yago}, and Wikidata~\cite{wikidata}.

Knowledge graph reasoning refers to the process of deriving new knowledge or insights from existing knowledge graphs in response to a query ~\cite{10.1145/3580305.3599564}. In essence, the knowledge graph reasoning pipeline comprises three key elements: the input query, background knowledge, and reasoning model, each posing unique challenges. The background knowledge may vary in completeness, influencing the system's ability to accurately interpret and utilize information. Meanwhile, input queries range from clear and specific to ambiguous or dynamically changing, demanding robust mechanisms for understanding user intent. Furthermore, the reasoning model's approach, whether inductive or deductive, impacts the system's ability to draw meaningful conclusions from the available data. Addressing these challenges necessitates adaptable algorithms and techniques to ensure the efficacy and reliability of knowledge graph reasoning across diverse contexts and applications.

Recently, there is a trend to utilize neural-symbolic artificial intelligence to enhance reasoning on knowledge graphs ~\cite{zhu2022neural}. Since knowledge graphs can be viewed as discrete symbolic representations of knowledge, it is natural to integrate knowledge graphs with neural models to unleash the full potential of neural-symbolic reasoning. Traditional symbolic reasoning is intolerant of ambiguous and noisy data, but knowledge graphs are often incomplete, which brings difficulties to symbolic reasoning. On the contrary, the recent advances in deep learning promote neural reasoning on knowledge graphs, which is robust to ambiguous and noisy data but lacks interpretability compared to symbolic reasoning. Considering the advantages and disadvantages of both methodologies, recent efforts have been made to combine the two reasoning methods for better reasoning on knowledge graphs.

Last but not least, the emergence of large language models (LLMs) ~\cite{gpt2,deepseekai2025deepseekr1incentivizingreasoningcapability} has shown impressive natural language capabilities. However, they struggle with logical reasoning that requires structured knowledge. The integration of LLMs with knowledge graphs during the reasoning process represents a promising area of exploration. While some methods have been proposed in this regard, a large part of this topic is unexplored or under-explored.

This survey provides a comprehensive exploration of knowledge graph reasoning for different types of queries. We introduce knowledge graph reasoning for single-hop queries, complex logical queries, and natural language queries. Furthermore, the integration of neural symbolic artificial intelligence techniques is investigated, highlighting innovative methodologies for enhancing reasoning capabilities within knowledge graphs. Lastly, the survey delves into the fusion of Large Language Models (LLMs) with knowledge graph reasoning, showcasing advancements at the intersection of natural language processing and structured data reasoning.

While numerous surveys have explored knowledge graph embedding, few have explicitly addressed knowledge graph reasoning from both query types and neural symbolic perspectives. This paper aims to fill this gap by introducing different topics and offering a comprehensive overview of knowledge graph reasoning from diverse viewpoints. Through detailed classification and elaboration within each category, this paper provides a holistic understanding of the complexities and advancements in knowledge graph reasoning, bridging the gap between different query types and neural symbolic reasoning, and offering insights into future directions for this field.

The remainder of the article is structured as follows. Section \ref{preliminary} provides an overview of the background knowledge closely associated with knowledge graph reasoning and neural symbolic reasoning. Section \ref{single} delves into knowledge graph reasoning for single-hop queries, examining various perspectives. Following this, Section \ref{complex} explores the intricacies of reasoning with complex logical queries. In Section \ref{nlp}, we scrutinize knowledge graph reasoning for both single-turn and multi-turn queries in natural language. Finally, we draw conclusions based on the insights gained from our survey.

%% file: 002preliminary.tex
In this section, we first formally define knowledge graphs and knowledge graph reasoning.

\subsection{Definition and Notation}

Knowledge graph reasoning refers to the process of using a knowledge graph, a structured representation of knowledge, as the basis for making logical inferences and drawing conclusions. 
More formally, the research question can be defined as

\begin{definition}
(\textit{Knowledge Graph})
Let \( G = (V, E, R) \) be a knowledge graph, where
\( V \) is the set of entities,
\( E \) is the set of relationships,
\( R \) is the set of triples \( (v_i, r_j, v_k) \) denoting relationships between entities, where \( v_i, v_k \in V \) and \( r_j \in E \).
\textbf{Knowledge graph reasoning}: Answer queries by traversing and reasoning over the graph.
\end{definition}

\begin{definition}
(\textit{Knowledge Graph Reasoning})
Let \( G = (V, E, R) \) be a knowledge graph, where
\( V \) is the set of entities,
\( E \) is the set of relationships,
\( R \) is the set of triples \( (v_i, r_j, v_k) \) denoting relationships between entities, where \( v_i, v_k \in V \) and \( r_j \in E \).
\textbf{Knowledge graph reasoning}: Answer queries by traversing and reasoning over the graph.
\end{definition}

\subsection{Symbolic Reasoning }

Symbolic reasoning in knowledge graphs refers to the process of deriving logical conclusions and making inferences based on symbolic representations of entities, relationships, and rules within the graph structure ~\cite{survey}. In this context, symbols represent entities or concepts, while relationships denote connections or associations between them. Symbolic reasoning involves applying logical rules ~\cite{rnnlogic} and operations to manipulate these symbols, enabling the system to perform tasks such as deductive reasoning ~\cite{survey2}, semantic inference, and knowledge integration. By leveraging symbolic representations and logical reasoning, knowledge graphs can facilitate complex problem-solving, semantic understanding, and decision-making in various domains, ranging from natural language processing to artificial intelligence applications.

\begin{figure*}
	\centering
	\includegraphics[width=0.99\textwidth]{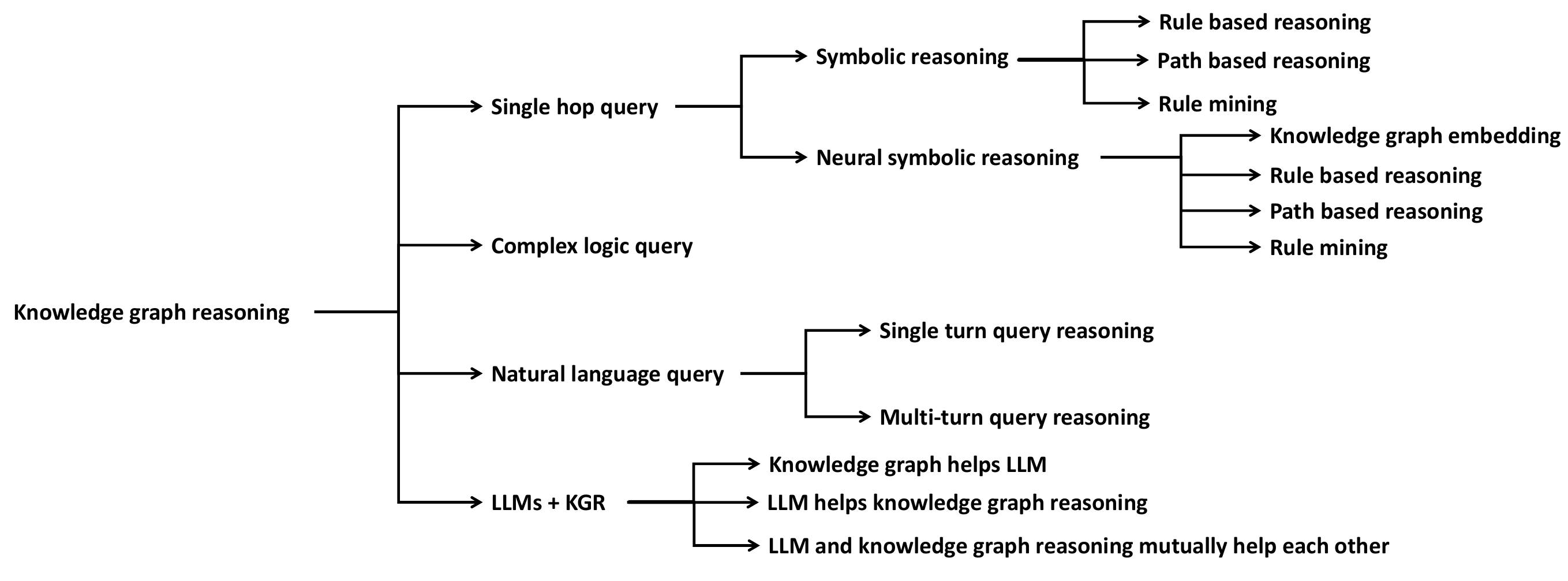}
	\vspace{-1\baselineskip}
	\caption{Survey framework.}
	\label{survey_framework}
\end{figure*}

\subsection{Neural Reasoning }

Neural reasoning in knowledge graphs refers to the utilization of neural network models to perform reasoning tasks directly on the graph structure ~\cite{survey}. Unlike traditional symbolic reasoning approaches, which rely on explicit rules and logical operations, neural reasoning leverages the power of deep learning techniques to learn implicit patterns and relationships within the graph. Through the use of neural networks, knowledge graphs can capture complex, non-linear dependencies between entities and infer higher-level knowledge from the graph's interconnected nodes. Neural reasoning methods often involve embedding entities and relationships into continuous vector spaces ~\cite{rotate,bordes2013translating,complex}, allowing neural networks to efficiently process and reason over large-scale knowledge graphs. This approach enables knowledge graphs to handle uncertainty, noise, and incompleteness in the data, making neural reasoning a promising paradigm for various applications, including question answering, recommendation systems, and semantic understanding.

\subsection{Neural Symbolic Reasoning }

Neural symbolic reasoning represents a fusion of neural network-based approaches with symbolic reasoning techniques, aiming to leverage the strengths of both paradigms in handling complex reasoning tasks ~\cite{survey}. In this framework, neural networks are used to learn representations of symbolic entities and relationships within a knowledge graph, capturing both their semantic meanings and structural dependencies. These learned representations are then combined with symbolic reasoning mechanisms to perform logical inference and reasoning tasks. By integrating neural and symbolic components, neural symbolic reasoning approaches strive to overcome the limitations of each individual approach. Neural networks offer the ability to learn from data and handle uncertainty, while symbolic reasoning provides formal logic-based reasoning and interpretability. This hybrid approach has shown promise in various domains, including natural language understanding ~\cite{embedkgqa}, knowledge graph reasoning ~\cite{bordes2013translating}, and automated theorem proving, by enabling more robust and flexible reasoning capabilities.

\subsection{Deductive Reasoning }
Knowledge graph deductive reasoning is a method used to derive new information from existing data within a knowledge graph by applying logical rules ~\cite{survey2}. A knowledge graph structures information as a network of entities and their interrelationships, represented in triples of subject-predicate-object. Deductive reasoning in this context involves using established logical rules to infer new facts ~\cite{survey2}. For instance, if the knowledge graph contains the triples "Alice works at XYZ Corp" and "XYZ Corp is located in New York," a rule might deduce that "Alice works in New York." This process leverages the structured nature of the graph and the logical relationships between entities to expand the knowledge base, ensuring that new conclusions are logically consistent with the existing data.

\subsection{Inductive Reasoning }

Knowledge graph inductive reasoning involves deriving generalized conclusions from specific instances within a knowledge graph ~\cite{survey2}. Unlike deductive reasoning, which applies general rules to specific cases, inductive reasoning identifies patterns and regularities in the data to formulate broader generalizations or hypotheses. For example, if a knowledge graph contains numerous instances where employees of various companies in New York tend to have a high turnover rate, inductive reasoning might lead to the hypothesis that companies in New York generally experience higher employee turnover. This approach allows for the generation of new insights and predictive models by examining trends and correlations in the data, providing a foundation for further exploration and hypothesis testing within the structured framework of a knowledge graph.

\subsection{Abductive Reasoning }
Knowledge graph abductive reasoning is a process used to infer the most likely explanation for a given set of observations within a knowledge graph ~\cite{survey2}. This type of reasoning seeks to find the best hypothesis that explains the observed data, often dealing with incomplete or uncertain information. For example, if a knowledge graph shows that a person has visited multiple cities known for tech conferences, abductive reasoning might infer that the person is likely involved in the tech industry. Unlike deductive reasoning, which guarantees the truth of the conclusion if the premises are true, or inductive reasoning, which generalizes from specific instances, abductive reasoning focuses on finding the most plausible explanation. This method is particularly useful in situations where there are multiple possible explanations, and it aims to identify the one that best fits the available evidence within the structured relationships of a knowledge graph.

\subsection{Paper Organization}

In this section, we've laid the groundwork by defining knowledge graph reasoning and discussing the related background knowledge. 
Moving forward, we'll delve into three distinct perspectives: reasoning for single-hop queries, complex logical queries, and natural language questions. Each perspective offers valuable insights into how knowledge graph reasoning operates and evolves in different contexts. By examining these perspectives, we aim to provide a comprehensive understanding of the diverse challenges and advancements within the field of knowledge graph reasoning.The taxonomy of this paper is illustrated in Figure ~\ref{survey_framework}.

%% file: 003single.tex
Reasoning for single-hop queries is a common task in the field of knowledge graphs, often referred to as knowledge graph completion. The objective here is to predict the tail entity $t$ given the head entity $h$ and the relation $r$, or conversely, to predict the head entity $h$ given the tail entity $t$ and the relation $r$.
In addition to entity prediction, there is the relation prediction task, where the goal is to predict the relationship $r$ between a given head entity $h$ and a tail entity $t$. This task can also be considered a specialized form of single-hop query.

A variety of methods have been proposed to address these tasks. In this section, we categorize the different approaches into three main types: symbolic, neural, and neural-symbolic methods, which will be elaborated in the following subsections.

\subsection{Symbolic Methods}
\subsubsection{Hard symbolic rule based reasoning.}
Symbolic rule reasoning methods rely on logical reasoning and explicit rules within the knowledge graph. They often utilize  rule-based or path-based inference techniques to deduce the missing entity or relation. Examples include rule mining algorithms and path ranking methods and so on.

\begin{figure}[h]
	\centering
	\includegraphics[width=0.4\textwidth]{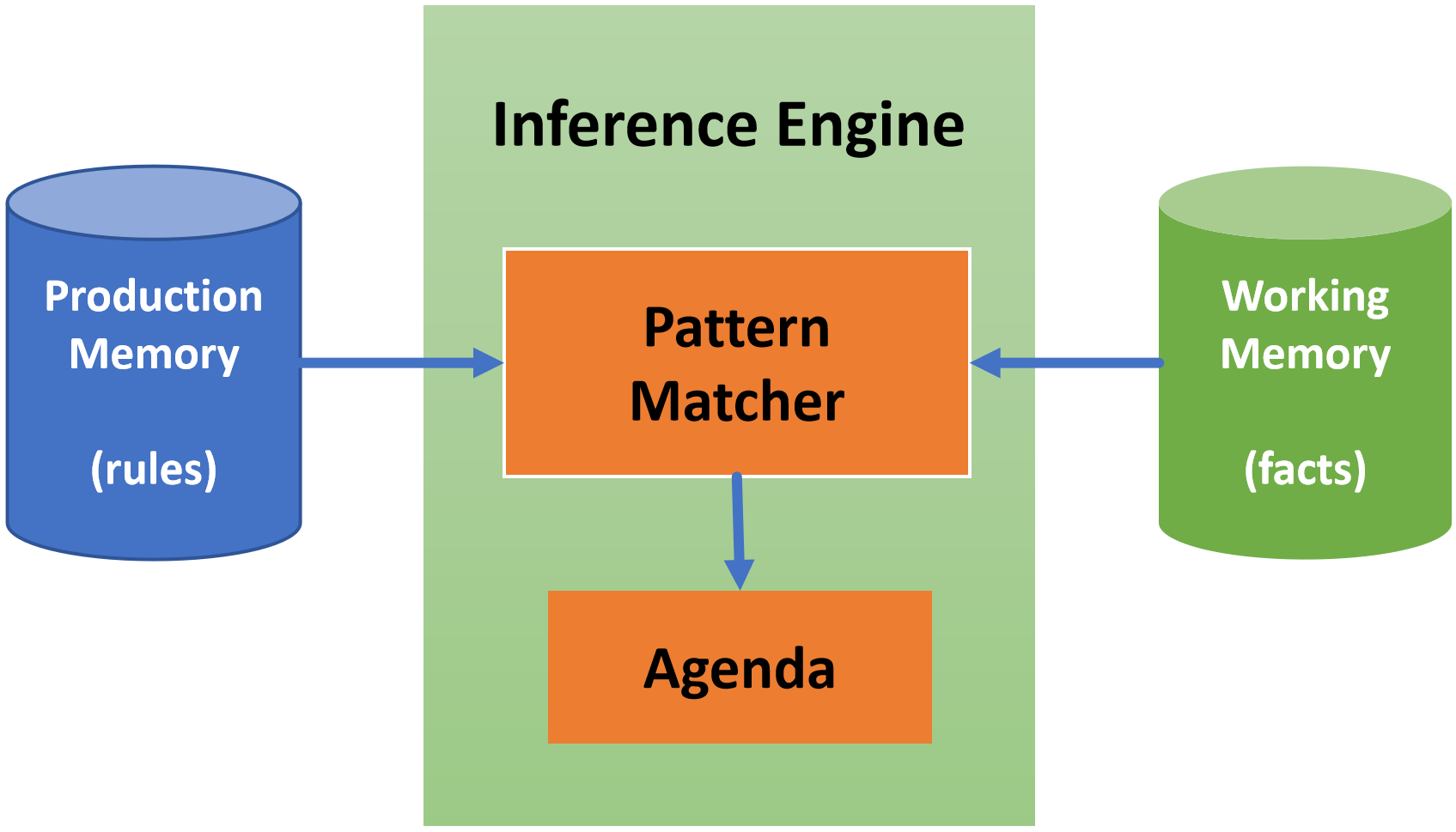}
	\caption{Rule based expert system. Sourced from ~\cite{10.1145/3580305.3599564}.}
	\label{rule_expert_system}
	\vspace{-1\baselineskip}
\end{figure}

One of the earliest symbolic rule reasoning methods can be track back to 1970s, which is the rule-based expert system ~\cite{abraham2005rule} as shown in Figure ~\ref{rule_expert_system}. The key idea of rule-based expert system is to apply hard rules iteratively to generate new facts. It is a very straightforward method. Generally, there are three primary components in ruled-based expert system: the inference engine, the knowledge base and the rules defined by experts. The inference engine can be treated as the brain of the reasoning system. It utilizes two methods to infer new knowledge according to the rules. The first method is called forward chaining. The idea is to start with facts and repeatedly apply the given rules to generate new facts, until no more rules can be applied. The second method is called backward chaining. It is goal oriented, where a goal usually refers to the query from the users. The backward chaining approach will apply the given rules by matching the goal to infer the answer.  Besides rule-based expert system,

 Alongside rule-based expert systems, Prolog ~\cite{clocksin2003programming} serves as a logic programming language adept at symbolic reasoning through facts and rules, commonly employed in AI applications for knowledge graph querying and manipulation. Similarly, Datalog ~\cite{ceri1989you} functions as a declarative logic programming language extensively used for knowledge graph reasoning. It excels in representing complex relationships and hierarchical structures succinctly, enabling tasks such as inference and consistency checking within knowledge graphs. Its logical foundations offer a robust framework for extracting meaningful insights from interconnected data.

Lastly, OWL (Web Ontology Language) ~\cite{mcguinness2004owl} is a formal language for defining and sharing ontologies within knowledge graphs on the Semantic Web. OWL enables detailed descriptions of entity classes, properties, and relationships, supporting automated reasoning to ensure data consistency and classification. By providing a standard framework, OWL facilitates interoperability and data integration across systems, enhancing the semantic richness and utility of knowledge graphs.

\subsubsection{Soft symbolic rule based reasoning.}
Despite the idea of hard rule-based reasoning is quite intuitive, it’s not the best solution most of the time. Because it requires experts to build rules based on their past experiences and intuitions. So, sometimes, mistakes may be made. Besides, the reasoning method can be very slow because it adds new facts to the knowledge base repeatedly. More importantly, it is not suitable for real time applications. It also lacks flexibility. For example, The reason process will give a world zero probability even if it only violates one formula. But this is not the real-world case. On example is “smoking causes cancer and friends have similar smoking habits”. These two rules are not always true, because not everyone who smokes gets cancer. And not all friends have similar smoking habits. And soft rules are more useful in this case because if a word violates a formula, it becomes less probable, not impossible.

\begin{figure}[h]
	\centering
	\includegraphics[width=0.45\textwidth]{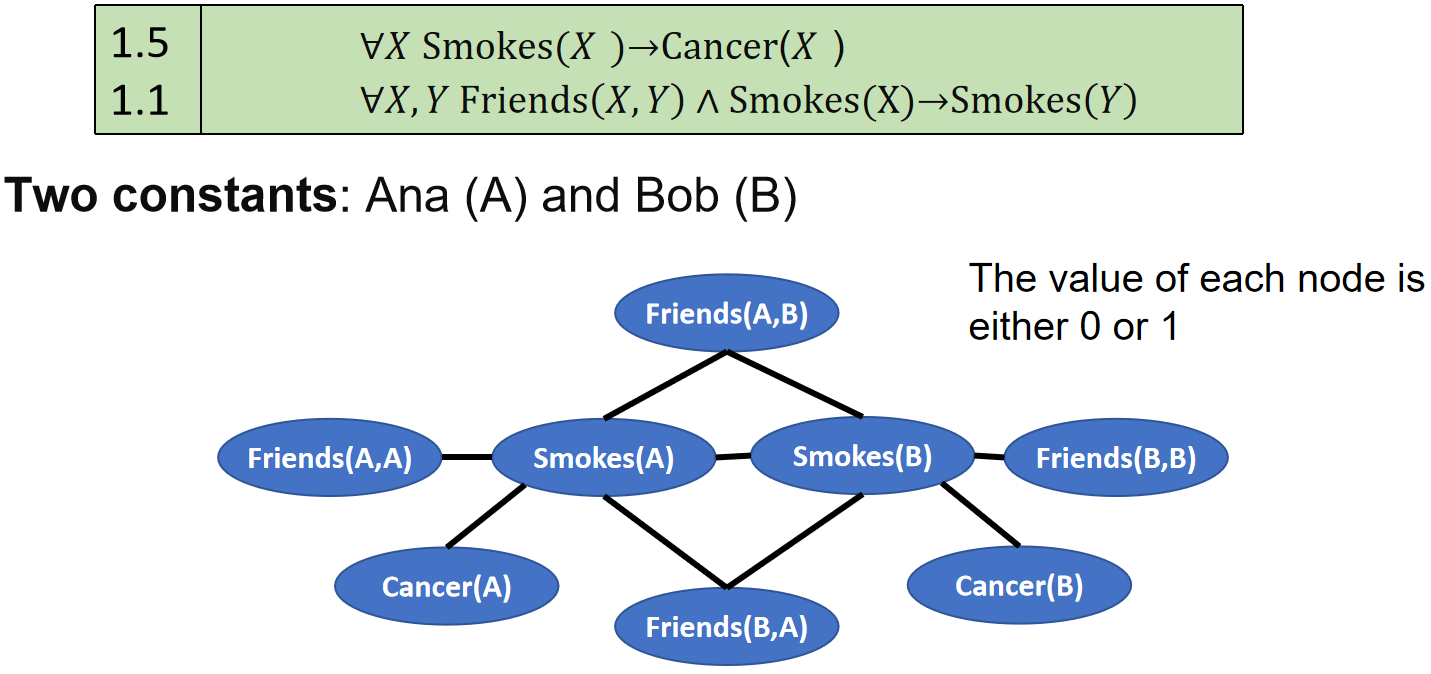}
	\caption{Example of markov logic network. Sourced from ~\cite{richardson2006markov}.}
	\label{markov_logic_network}
	\vspace{-1\baselineskip}
\end{figure}

Symbolic reasoning methods like Markov logic network ~\cite{richardson2006markov} is a representitive work to reason based on soft rules. The intuition of Markov Logic Network is to give Each formula an associated weight to reflect how strong a constraint it is. If the weight is higher, it’s a strong constraint, otherwise, it’s a weak constraint. When the weights go to infinite, it becomes hard rule reasoning. Formally, Markov network is an undirected graphical model with Node represent variables. And a potential function is defined for each clique in the graph. The distribution of the Markov network is defined as the multiplication of all the potential functions. When applying Markov network to soft rule reasoning, the idea is to treat the first order logic rules as the templates of Markov networ, and reason according to their weights. 
For example, “smoking causes cancer and friends have similar smoking habits” are two formulas, and they have weight 1.5 and 1.1 respectively, as shown in Figure ~\ref{markov_logic_network}.

Different from Markov logic networks which repeatedly use existing rules to infer knowledge, TensorLog ~\cite{cohen2016tensorlog} aims to find answers for a given query by matrix multiplication. 
The idea of Tensorlog is to formulate the reasoning process as a function and represent each entity as a one-hot vector. When applying the function to the input vector, the result is an n by 1 vector where the ith element denotes the probability that entity i is the answer.

\begin{figure}[h]
	\centering
	\includegraphics[width=0.45\textwidth]{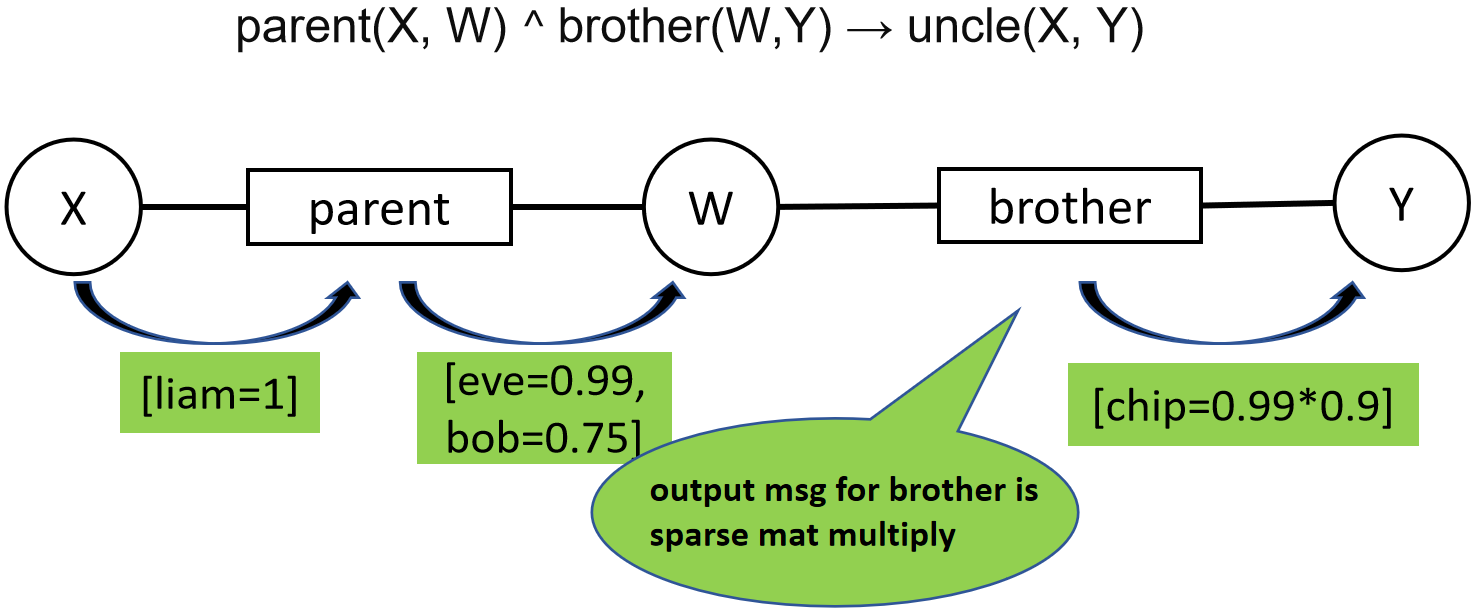}
	\caption{Example of Tensorlog.  Sourced from ~\cite{cohen2016tensorlog}}
	\label{tensorlog}
	\vspace{-1\baselineskip}
\end{figure}

An example of Tensorlog is inferring family relations like ``uncle'' in Figure ~\ref{tensorlog}. Given the logic rule \(\text{parent}(X, W) \land \text{brother}(W, Y) \rightarrow \text{uncle}(X, Y)\), Tensorlog takes the vector of \(X\) and multiplies it by the matrix of \(\text{parent}\). Then we get \(W\). We multiply the matrix of \(\text{brother}\) and get the output \(Y\). When the length of the rule becomes longer, and if there are multiple rules, the result is the summation of all the matrix multiplication sequences, and each matrix multiplication sequence has a weight \(a\).

\subsubsection{symbolic path based reasoning.}
For the rule based reasoning methods, they require the rule to be given. However, it’s not the case usually. We don’t have any rules. One possible solution is path-based reasoning. For the path based reasoning method, no rule is needed. It utilizes different paths in the knowledge graph to infer new information. These paths can be directed or undirected.

The very first method of path reasoning is path ranking algorithm ~\cite{lao2011random} which treats the random walks around a query node as the relational features. The idea of PRA is to use random walk to find many different paths between two nodes and treat these paths as the feature of the relation, and use a logistic regression model to learn a classifier to predict the truthfulness of the triplet.

After PRA, ProPPR ~\cite{gardner-etal-2014-incorporating} incorporates vector similarity into random walk inference over KGs to mitigate the feature sparsity inherent in using surface text. Specifically, when conducting a series of edge traversals in a random walk, ProPPR allows the walk to explore edges that exhibit semantic similarity to the given edge types, as defined by vector space embeddings of the edge types. This integration of distributional similarity and symbolic logical inference aims to alleviate the sparsity of the feature space constructed by PRA.

\subsubsection{symbolic rule mining.} Rule mining can also be treated as a special type of single hop query answering. Instead of directly answer which entity might be the correct answer, Rule mining aims at deducing general logic rules from the knowledge graphs. The entities derived from the given head entity and the query relation following the logic rules are returned as the answers.

AMIE ~\cite{galarraga2013amie} delves into logic rule exploration through a two-step process. Initially, it engages in Rule Extending, wherein candidate rules undergo expansion via three distinct operations: Add Dangling Atom, Add Instantiated Atom and Add Closing Atom. Subsequently, in the Rule Pruning phase, it sifts through the rules, discarding those deemed faulty, and selects the confident ones based on predefined evaluation metrics. In terms of implementation, the approach leverages SPARQL queries on graph databases to sift through the knowledge graphs (KGs), identifying suitable facts $(h, r, t)$ that adhere to the extended rules from the first step and surpass the specified metric thresholds from the second step.

After AMIE, the subsequent algorithm, AMIE+ \cite{galarraga2015fast}, enhances the efficiency of AMIE's implementation through adjustments to both the Rule Extending process and the evaluation metrics in the Rule Pruning phase. In the Rule Extending stage, AMIE+ selectively extends a rule only if it can be completed before reaching the predefined maximum rule length. To elaborate, it refrains from appending dangling atoms in the final step, which would introduce fresh variables and lead to non-closure. Instead, AMIE+ waits until the last step to incorporate instantiated atoms and closing atoms, thereby ensuring rule closure. Additionally, the SPARQL queries employed for rule search are streamlined. For instance, when appending a dangling atom to a parent rule $R_p$ to generate a child rule $R_c$, if the predicate of the new atom already exists in $R_p$, the support for $R_c$ remains the same as that of $R_p$. Consequently, recalculating support for $R_c$ becomes unnecessary, thus expediting the SPARQL query process.

While AMIE and AMIE+ have found extensive use across various scenarios, they still face scalability challenges when dealing with large knowledge graphs (KGs). This limitation stems from their reliance on projection queries executed via SQL or SPARQL, where reducing the vast search space remains challenging. In response, RLvLR \cite{omran2018scalable} employs an embedding technique to sample relevant entities and facts pertaining to the target predicate/relation, significantly curtailing the search space. Firstly, RLvLR samples a sub-knowledge graph relevant to the target predicate. Secondly, it utilizes the RESCAL knowledge graph embedding model \cite{nickel2011three} to generate embeddings for entities and relations in the subgraph, with the embedding for a predicate augmentation being the average value of associated entity embeddings. Thirdly, RLvLR employs a scoring function based on these embeddings to guide and prune rule search, proving effective in rule extraction. Finally, candidate rules are evaluated based on metrics such as $hc$ and $conf$, akin to AMIE, computed efficiently through matrix multiplication. By incorporating the embedding technique, RLvLR significantly enhances the efficiency of the rule search process. Another method RuLES \cite{ho2018rule} utilizes the embedding technique to assess the quality of learned rules. It incorporates external text information of entities to derive their embeddings, enabling the calculation of confidence scores for facts. RuLES defines the external quality of a learned rule as the average confidence score of all derived facts. Ultimately, RuLES integrates statistical and embedding measures to more precisely evaluate the quality of learned rules.

\subsubsection{Summary}
In this section, we explore various methods relevant to symbolic reasoning. We discuss rule-based approaches, which leverage logical rules for inference, as well as path-based methods, which analyze patterns within knowledge graphs. Additionally, we delve into rule mining techniques, which aim to extract logical rules from structured data sources. Each method offers unique insights and capabilities in the realm of symbolic reasoning.

\subsection{Neural-Symbolic Methods}

Neural-symbolic methods aim to combine the strengths of both symbolic and neural methods. They often incorporate symbolic reasoning within a neural framework or use neural networks to enhance symbolic inference.

\subsubsection{Knowledge graph embedding}

Knowledge graph embedding usually encodes entities as low-dimensional vectors and encodes relations as parametric algebraic operations in the continuous space. The basic idea is to design a score function \( f \) which takes the triplet embedding as input, so that a true triplet receives a higher score than a false one. 
There are a lot of applications which utilize knowledge graph embedding. One of the most common applications is knowledge graph completion. For example, given a head entity and a tail entity, the missing relation is the one which maximizes the score function value. Likewise, given a head entity and a relation, the missing tail entity is the one which, again, maximizes the score function value.

Many KG embedding methods have been developed. The basic idea of TransE ~\cite{bordes2013translating} is to view the relation \( r \) as the transition from the head entity to the tail entity. Mathematically, it means that ideally, the tail entity \( t \) should be the summation of the head entity and the relation. 
Another method is DistMult ~\cite{DistMult}. Similar as TransE, DistMult also embed entities and relations into vectors in the real/encludience space. 
Different from TransE, DistMult views the relation $r$ as the elementwise weights of the head entity $h$.
Its score function is defined as the weighted sum over all elements of the head entity by the corresponding elements in the relation.
So, in DistMult, the ideal tail entity should be $h\dot r$.
Another method ComplEx ~\cite{complex} embeds entities and relations in Complex vector space. Each embedding now has a real part and an imaginary part. 
Given a point $z$ which is $x + iy$ in the embedding space, its conjugate $\bar{Z}$ is $x - iy$. 
The scoring function used in complex is very similar to that of distmult. But 
We replace $t$ by its conjugate we only taking the real part of the function. 
Different from dot product, in Complex ~\cite{complex} Hermitian dot product $<h, r, \bar{t}>$ is asymmetric, where $\bar{t}$ is the complex conjugate of $t$. Thus it naturally captures the anti-symmetry.
Another method is called RotatE ~\cite{rotate}. The key idea of rotatE is to solve the limitions in previous methods. similar to complex, rotatE also represent head and tail entities and relation in complex vector space,
Different from complex, all the relation embedding are modelled as rotation from the head entity $h$ to the tail entity $t$. Compared with other methods, RotatE can support different relation properties, such as symmetry/antisymmetry, inversion, and composition. 
Other methods such as TransH \cite{wang2014knowledge} embeds knowledge graph into the hyperplane of a specific relationship to measure the distance. TransR \cite{lin2015learning} represents entities and relationships in separate entity and relationship spaces. 
The semantic matching model uses semantic similarity to score the relationship between head entities and tail entities. RESCAL \cite{nickel2011three} treats each entity as a vector to capture its implied semantics and uses the relationship matrix to model the interactions between latent factors. QuatE \cite{QuatE} uses two rotating planes to model the relations to a hyper-complex space. HolE \cite{HolE} employs cyclic correlation to represent the composition of the graph. However, neither of these methods captures the structure information of the graph which should be important to the graph.

In addition to point embeddings, recent methods have explored using geometric regions to represent knowledge graphs in embedding spaces. Geometric embedding techniques include regions like boxes and spheres, which are effective in modeling relationships and hierarchical structures within knowledge graphs ~\cite{probabilistic_2018_ACL_Luke, representing_words}.
Other approaches employ probabilistic embeddings to represent knowledge graphs. For instance, probability-based embeddings, such as Gaussian distributions, capture uncertainty and variability in the data, providing a probabilistic interpretation of the embeddings ~\cite{gaussian_embedding, vilnis2015word}. These methods enhance the expressiveness and flexibility of embeddings, enabling more robust reasoning and inference in various applications.

\subsubsection{Neural symbolic rule based reasoning}
Neural LP ~\cite{yang2017differentiable}, which is a generalization of Tensorlog that focuses on learning logical rules with confidence scores. In Tensorlog, the reasoning process is a sequence of matrix multiplication operations. Tensorlog denotes the input query entity as a one-hot vector and each relation as a matrix \( R \). The reasoning results are computed by matrix multiplication, retrieving entities whose entries are non-zero as answers.
Neural LP adopts the same idea as Tensorlog. In Neural LP, the authors found that when the rule length increases from 2 to \( L \), the original first matrix multiplication then summation process can be rewritten as the first summation then multiplication process. After changing the order of the operations, the original matrix multiplication process becomes learning the combination of relationships at each step. This process can be modeled by a recurrent neural network (RNN) for \( T \) steps. 
The candidate pool of Neural LP is very large, which leads to a huge search space. So, it’s hard to identify useful rules in the search space. Most of the time the weights may not reflect the importance of rules precisely. To solve these limitations, RNNLogic ~\cite{rnnlogic} treats all logic rules as latent variables. That is, to answer a query, there may be more than 10 or 20 related rules, and we treat all these logic rules as latent variables. In this way, the rule mining problem becomes a rule inference problem. RNNLogic contains two components: the rule generator, which will generate some candidate logic rules for a specific query, and a reasoning predictor, which is used to predict how likely we can find the answer given a logic rule. Different from Neural LP, the search space of RNNLogic is much smaller. Because all logic rules are treated as latent variables, the EM algorithm can be used for inference.

\subsubsection{Neural symbolic path based methods}
Previously, we introduced symbolic path-based reasoning methods. Neural symbolic path-based methods aim to answer single-hop queries by combining neural and symbolic techniques, utilizing path information for more robust reasoning.

Existing symbolic path ranking methods consider only the direct path information between two entities, neglecting the rich context information surrounding entities. This often leads to suboptimal solutions. To address this, PathCon ~\cite{pathcon} incorporates both relational context and relational paths in the reasoning process. Relational context refers to the k-hop neighboring relations of a given entity, while relational paths are the connections between two entities. PathCon encodes relational context using Relational Message Passing to aggregate k-hop content information around a predicate. For encoding relational paths, PathCon identifies all paths between entities $h$ and $t$ of length no greater than k, and then uses an RNN to learn the representation of each path.
After learning both context and path information, PathCon combines them. It concatenates the final embeddings of entities $h$ and $t$ and processes them through a neural network to obtain the final context embedding. An attention mechanism aggregates the relational path information. The final output is a probability distribution, computed based on the combined context and path embeddings.

Unlike previous methods, DeepPath ~\cite{deeppath} uses reinforcement learning to predict missing links. DeepPath learns paths rather than relying solely on random walks, framing the path-finding process as a Markov decision process. It trains a reinforcement learning agent to discover paths, using these paths as horn clauses for knowledge graph reasoning. In DeepPath, the agent is represented by a neural network, and the answer-finding process is modeled as a Markov decision process. The states are defined as the concatenation of the current entity embedding and the distance between the target and the current entity in the embedding space, while the action space consists of all unique relation types in the knowledge graph.

\subsubsection{Neural symbolic rule mining}

For all previous symbolic rule mining methods, the focus is on mining Horn clauses. In GraIL~\cite{grail}, the authors propose using subgraphs for reasoning, based on the idea that useful rules are contained in the subgraph around the query. GraIL applies graph neural networks (GNNs) to subgraphs surrounding the candidate edge, hypothesizing that subgraph structure provides evidence for inferring relations between nodes. This GNN-based method avoids explicit rule induction while remaining expressive enough to capture logical rules.
The reasoning process in GraIL comprises three steps: first, extracting a subgraph around the candidate edge; second, assigning structural labels to nodes based on their distances from the target nodes to encode graph substructure; and third, running a GNN on the extracted subgraph to capture the rules. The GNN in GraIL uses the traditional combine-and-aggregate paradigm, where each node aggregates information from its neighbors using an aggregate function. To distinguish different relation types in the knowledge graph, GraIL employs relation-specific attention to weigh information from different neighbors.
During inference, given a triplet \( (u, r_t, v) \), GraIL obtains the subgraph representation through average pooling of all latent node representations. It then concatenates this subgraph representation with the target nodes’ latent representations and a learned embedding of the target relation. These concatenated representations are passed through a linear layer to obtain the score. The model is trained using gradient descent.

\subsubsection{Summary}

Neural-symbolic reasoning combines the strengths of neural networks and symbolic reasoning to tackle the complexities of knowledge graphs. Traditional symbolic reasoning methods, like rule-based expert systems, employ predefined rules for inference, while path-based approaches like the Path Ranking Algorithm utilize random walks to infer relationships. Hybrid methodologies, such as PathCon, integrate relational context and path information through neural networks, enhancing reasoning capabilities. Embedding techniques, including geometric and probabilistic embeddings, represent entities and relationships in continuous vector spaces, facilitating more flexible knowledge graph operations. Reinforcement learning-based methods, like DeepPath, utilize trained agents to navigate knowledge graphs and predict missing links. By merging neural and symbolic techniques, neural-symbolic reasoning offers a comprehensive approach to understanding and reasoning over complex knowledge graph structures, promising advancements in various applications requiring automated reasoning and inference.

%% file: 005complex.tex
In this section, we generalize the query into logically more complex forms~\cite{marker2006model} and explain how to solve them using neural and symbolic methods.
Compared to simple-hop queries, the additional complexity is introduced by involving more ``elements'' in logical language, such as multiple predicates, quantifiers, and variables.
The fundamental motivation of complex queries also follows the narrative of single-hop query prediction, where we want to derive new knowledge but with more logical constraints. We also refer readers to the earlier survey~\cite{wang2022logical} in this direction. Both single-hop queries and multi-hop queries fall under the broader category of complex queries. However, to differentiate knowledge graph completion from complex logical query answering, we treat single-hop queries and complex logical queries as two distinct components.

\subsection{What is Complex Query?}
General logical queries follow the definitions of mathematical logic and model theory~\cite{marker2006model}. The previous but perhaps not up-to-date survey provides more rigorous definitions and discussions~\cite{wang2022logical}. Regarding logical language coverage, complex queries studied on knowledge graphs are still preliminary compared to parallel studies in databases~\cite{libkin2004elements} and semantic web research~\cite{mcguinness2004owl}.

In the literature, a general term describing complex queries on knowledge graphs is the general ``first-order query'' or ``first-order logical query''~\cite{query2box,betaE}. However, recent rigorous characterization~\cite{yin2024rethinking} distinguished the queries discussed in the definition, and queries studied in empirical methods and benchmarks are two overlapping query families. The first is existential first-order queries that appeared in definitions of many works~\cite{betaE,newlook}, and the second is the tree-formed queries widely adopted in the empirical construction of benchmarks~\cite{wang2022benchmarking}. Most empirical results remain credible despite the fine-grained differences in query families.

We hereby introduce the definitions of two queries based on a fragment of the first-order language. A term is either an entity $e\in V$ or a variable. A variable is a non-determined entity whose value can be taken from $V$. A variable can be either existentially quantified or not. Universally quantified variables are not considered yet in the literature. An atomic formula is $r(a, b)$ where $r\in E$ is the relation. Then, we define the Existential First-Order (EFO) formulae.

\begin{definition}[Existential First Order Formula]\label{def:formula}
The set of the existential formulas is the smallest set $\Phi$ that satisfies the following property:
\begin{compactitem}
    \item[(i)] For atomic formula $r(a,b)$, itself and its negation $r(a,b), \lnot r(a,b)\in \Phi$, where $a,b$ are either entities in $V$ or variables, $r$ is the relation in $E$.
    \item[(ii)] If $\phi, \psi\in \Phi$, then $(\phi\land \psi),(\phi \lor \psi) \in \Phi$
    \item[(iii)] If $\phi \in \Phi$ and $x_i$ is any variable, then $\exists x_i \phi \in \Phi$.
\end{compactitem}
\end{definition}

When there is more than one free variable, an EFO formula is an EFO query. In most previous studies, only one existential variable is considered, leading to the family of EFO-1, denoted as $\Phi$. The families with more than one variable are titled EFO-k~\cite{yin2024efo}; so far, there is no specific method targeting EFO-k. The key feature of EFO queries is that the logical negation is only attached to atomic formulas, defined by rule (i). Consequently, one can always move existential quantifiers at the beginning of the formula as the prenex normal form~\cite{marker2006model}. Moreover, it is always convenient to reorganize the logical conjunctions and disjunctions into either Disjunctive Normal Form (DNF) or Conjunctive Normal Form (CNF). One common way to define the EFO-1 query is by the DNF and the conjunctive queries.

Specifically, the queries $q$ can be written as follows. 
\begin{definition}[EFO-1 Query]
An EFO-1 query is defined as
\begin{align}
    q(y) = \exists x_1, ..., x_k, \pi_1(y) \lor \cdots \lor \pi_n(y),
\end{align}
where $y$ is the variable to be queried, $x_1, ..., x_k$ are existentially quantified variables, and $\pi_n(y)$ is the conjunctive query to be defined in the following parts.
\end{definition}
\begin{definition}[Conjunctive Query]
A conjunctive logical query is written as
\begin{align*}
\pi_i(y) = \exists x_1, ..., x_k : \varrho^i_1 \land \varrho^i_2 \land ... \land \varrho^i_m,
\end{align*}
where each $\varrho^i_j$ is the atomic formula $r(a , b)$ or its negation $\lnot r(a, b)$.
\end{definition}

Another query family that is well studied is formally defined as the Tree-Formed (TF) queries $\Phi_{\rm tf}$.
\begin{definition}[Tree-Form Query]\label{def:TF query}
The set of the Tree-Form queries is the smallest set $\Phi_{\rm tf}$ such that: 
\begin{compactitem}
\item[(i)] If $\phi(y) = r(a,y)$, where $a\in E$, $r\in R$, then $\phi(y) \in \Phi_{\rm tf}$; 
\item[(ii)] If $\phi(y)\in \Phi_{\rm tf}, \lnot \phi(y) \in \Phi_{\rm tf}$; 
\item[(iii)] If $\phi(y), \psi(y)\in \Phi_{\rm tf}$, then $(\phi\land \psi)(y) \in \Phi_{\rm tf}$ $(\phi \lor \psi)(y)\in \Phi_{\rm tf}$;
\item[(iv)] If $\phi(y) \in \Phi_{\rm tf}$ and $y^{\prime}$ is any variable, then $\psi(y^{\prime})= \exists  y. r(y,y^{\prime})\land \phi(y) \in \Phi_{\rm tf}$.    
\end{compactitem}
\end{definition}
One key feature of $\Phi_{\rm tf}$ is that the answer set can be constructed recursively through set operations, such as union, intersection, and compliment. As specifically shown in Definition~\ref{def:TF query}, rule (ii) corresponds to the set complement against an implicit universe set; rule (iii) relates to the set intersection and union to logical conjunction and disjunction; and rule (iv) for set projection. Under the context of tree-form queries, we use logical connectives (conjunction, disjunction, and negation) and set operations (intersection, union, and complement) interchangeably.

EFO-1 and tree-form query families are different but not mutually exclusive (EFO-1 $\cap$ TF $\neq \emptyset$). There are also queries in the tree-form family but not in EFO-1 and vice versa. Detailed discussions of query syntax can be found in~\cite{yin2024rethinking}.
The follow-up part then explains neural and neural-symbolic methods for TF and EFO-1 queries.

\subsection{Neural Methods}

Neural methods conduct logical reasoning in a fixed-dimensional embedding space, where previous insights from knowledge graph embeddings can be applied. The methods for tree-formed queries and EFO-1 queries differ significantly in two ways. In short, methods targeting tree-formed queries emphasize the modeling of set operations~\cite{wang2022benchmarking}. Methods for the EFO-1 query stress the DNF formulation and the conjunctive query.

\subsubsection{Tree-form query}
The first attempt to solve a tree-form query begins with the logical conjunction, or more specifically; the final answer set can be derived by set projection and set intersection.
As the earliest example, GQE~\cite{GQE} embeds graph nodes in a low-dimensional space and represents set projection and intersection as neural operations in the embedding space. Consequently, terms in the query, including constant entities, existential variables, and free variables, can be represented or computed as the embedding. Then, the nearest neighbor search is used to find answers. The embeddings and neural models are trained on synthetic datasets by an end-to-end auto-differentiation.
Follow-up methods followed the key design principles: (1) represent the terms into low-dimensional embeddings; (2) set operations are modeled by differentiable operations in the embedding spaces; (3) identify the final answers by nearest neighbor search. Moreover, the supported set operations are expanded to set union (logical disjunction) and set complement (logical negation).

Notably, the set intersection, union, and negation provide some natural properties and intuitions. An example is the box-embedding space and various query embedding methods. Query2Box~\cite{query2box} proposes to model queries as boxes (i.e., hyper-rectangles), where a set of points inside the box corresponds to a set of answer entities of the query. Set intersections can be naturally represented as geometric intersections of boxes in the space. On the other hand, the set union cannot be modeled by the geometric union of boxes because the resulting shape is not a box. This issue can be indirectly addressed by transforming queries into a DNF.
Furthermore, NewLook~\cite{newlook} adopts a neural network to relearn the box embedding at each projection operation to mitigate the cascading error problem and also introduces a new logic operator, \emph{set difference}, so that the set compliment can be modeled by the equivalently. Besides the box-embedding space, other kinds of embedding spaces are also widely explored, including the space of convex cones~\cite{zhang2021cone}, parametric probabilistic distributions~\cite{betaE,yang2022gammae}, and vectors~\cite{fuzzyQE,wang2023wasserstein}.

\subsubsection{EFO-1 query}
This part only focuses on methods that are capable of solving queries that are EFO-1 but not tree-form. Such queries are characterized by the query graph of the sub-conjunctive query. Specifically, such a query can be represented as a simple (with multi-edges) or cyclic graph, which prohibits the perspective that regards the logical query as compositional operations~\cite{yin2024rethinking}. Instead, a more natural framework is to consider the atomic formulas or their negation in the conjunctive query as constraints in constraint satisfaction problems. One practical approach is to adopt the graph neural networks on the query graph. LMPNN injects the logical constraint in the message and connects the complex query to message-passing networks~\cite{wang2023logical}. More sophisticated neural architectures like transformers are investigated on the query graph~\cite{xu2023query2triple} and messages~\cite{zhang2024conditional}.

\subsection{Neural-Symbolic Methods}

The neural-symbolic methods for complex query answering integrate symbolic algorithms, which have been extensively studied in the database community. The neural part of such approaches, on the other hand, is less capable than that of neural approaches. In practice, the neural part of neural symbolic approaches is mainly chosen as link predictors or knowledge graph embeddings. Different from the previous discussions on the neural approach, neural symbolic approaches rely heavily on the symbolic algorithm; thus, they can solve a more extensive set of queries.

Almost all symbolic algorithms search for a proper assignment of variables $x_1=e_1, ..., x_k=e_k$, and $y=a$ to satisfy the logical constraints in queries. Combined with neural link predictors, the boolean value of satisfaction is turned into a continuous score or normalized into $[0, 1]$ as fuzzy truth values. The preliminary approach models the adjacent matrices of KG for each relation, with elements as the fuzzy truth value of triples. The details of how to normalize the output of the link predictor into fuzzy truth values vary in different methods. However, it does not change the nature of the problem as a search process. Several search strategies can be seamlessly applied to such a problem. For example, beam search realized for the acyclic query graph is known as CQD~\cite{CQD}, and search on acyclic query graph with additional backtracking is proposed as QTO~\cite{QTO}. The generalization from acyclic to cyclic and multi-edge query graphs is known as FIT~\cite{yin2024rethinking}. Many algorithmic results are also available to accelerate the algorithm, such as using a count-min sketch in EMQL~\cite{EMQL} to store the entity set for each query node and using vector similarity to find similar results during the search process.

Neural link predictors can be deeply engaged with search and not just materialized as adjacency matrices. CQD-CO relaxed the search problem from symbolic assignment spaces into neural embedding space, thus enabling gradient-based optimization with differentiable link predictions~\cite{CQD}. CQD-A, as a more advanced method, can adapt knowledge graph embeddings from the feedback of the training data~\cite{arakelyan2024adapting}. A similar but technically different approach is the GNN-QE, where the link predictor is not just embeddings but a graph neural network to be learned from the feedback of the search results~\cite{zhu2022neural}.

Recently, some approaches have attempted to combine large language models (LLMs) with neural-symbolic methods to address this problem. For instance, in~\cite{liu2024logicquerythoughtsguiding}, a framework is proposed that decomposes complex questions into multiple subquestions, which are then individually answered by LLMs. Simultaneously, neural-symbolic methods are applied to incomplete knowledge graphs. At each time step, the results from the LLMs and knowledge graph reasoning are integrated to produce a cohesive response.

%% file: 004language.tex
\subsection{Reasoning for Single-turn Query}
When the input query is a natural language sentence, existing methods can be divided into several categories, such as semantic parsing-based methods, information retrieval-based methods, and embedding-based methods.

For example, PullNet~\cite{pullnet} is information retrieval-based methods that retrieve a subgraph of candidate answers from the knowledge base to guide prediction. KV-Mem~\cite{keyvalue} and EmbedKGQA~\cite{embedkgqa} are embedding and deep learning-based methods that use deep learning networks to embed the question into a point in the embedding space and find answers according to a similarity function. In PrefNet ~\cite{prefnet}, a reranking based method is used to rerank all candidate answers to get better results. In semantic parsing-based methods, a general strategy to answer the question is to transform the question into a query graph and search for the answer according to the query graph. For example, in ~\cite{RnG-KBQA}, Xi et al. propose a model with candidate query graphs ranking and true query graph generation components. By iteratively updating these components, their performance improves. The query graph generated can then be used to search the KG. In ~\cite{binet}, Liu et al. propose a multi-task model to tackle KGQA and KGC simultaneously. They transform the multi-hop query into a path in the knowledge graph and use it to complete the knowledge graph, jointly training the model for both tasks.

Recently, reinforcement learning-based methods have been used to answer natural language questions on the knowledge graph. Zhang et al.~\cite{ZHANG2022102933} use a KG as the environment and propose an RL-based agent model to navigate the KG to find answers to input questions. Similarly, in~\cite{go_for_a_walk, LinRX2018_MultiHopKG, deeppath}, authors use RL models to find paths in the KG for answering input queries. Other studies, such as ~\cite{misu-etal-2012-reinforcement, xiaobing, alexa, gpt2, google_lamda}, integrate RL with other methods to create more human-like systems.

\subsection{Reasoning for Multi-turn Query}

Various approaches have been used to reason for multi-turn questions. For instance, Conquer~\cite{conquer} notices that whenever there is a reformulation, it is likely that the previous answer was wrong, and when there is a new intent, it’s more likely that the previous answer was correct. Based on this idea, Conquer treats reformulations as implicit feedback to train a reinforcement learning model. The idea is to position multiple reinforcement learning agents on relevant entities, allowing these agents to walk over the knowledge graph to answer in its neighborhood. 

Despite the common use of reinforcement learning in conversational question answering, it has many disadvantages. For example, the paths in the knowledge graph found by agents are often very similar, making them hard to distinguish. In this example, all these five paths lead to the answer entity, but they have different intermediate nodes. Besides, the reward is sparse, making the model hard to train. To solve these problems, PRALINE~\cite{praline} uses a contrastive representation learning approach to rank KG paths for retrieving the correct answers effectively. Extracted KG paths leading to correct answers are marked as positive, while others are negative. Contrastive learning is ideal since it allows us to identify positive KG paths and answer conversational questions. Besides path information, the entire dialog history, the verbalized answers, and domain information of the conversation are also used to help learn better path embeddings. Continuing along this trajectory, CornNet~\cite{cornnet} advocates for the utilization of language models to generate additional reformulations. This approach aids in enhancing the model's comprehension of original, concise, and potentially challenging questions, ultimately leading to improved performance.

In~\cite{ask_the_right_question}, the authors employed reinforcement learning to train an agent that reformulates input questions to aid the system's understanding. In~\cite{Dialog-to-Action}, an encoder-decoder model is used to transform natural language questions into logical queries for finding answers. In~\cite{kacupaj-etal-2021-conversational}, a Transformer model is used to generate logical forms, and graph attention is introduced to identify entities in the query context. Other systems, such as Google's Lambda~\cite{google_lamda}, Amazon Alexa~\cite{alexa}, Apple's Siri, and OpenAI's ChatGPT, are also pursuing this task.

Question rewriting, which aims to reformulate an input question into a more salient representation, is also used in multi-turn question answering. This can improve the accuracy of search engine results or make a question more understandable for a natural language processing (NLP) system. In~\cite{qa_rewrite}, a unidirectional Transformer decoder is proposed to automatically rewrite a user's input question to improve the performance of a conversational question answering system. In~\cite{elgohary-etal-2019-unpack}, authors propose a Seq2Seq model to rewrite the current question according to the conversational history and introduce a new dataset named CANARD. In~\cite{10.1145/2623330.2623677}, query rewriting rules are mined from a background KG, and a query rewriting operator is used to generate a new question.

%% file: 006other.tex
\begin{figure*}
	\centering
	\includegraphics[width=0.8\textwidth]{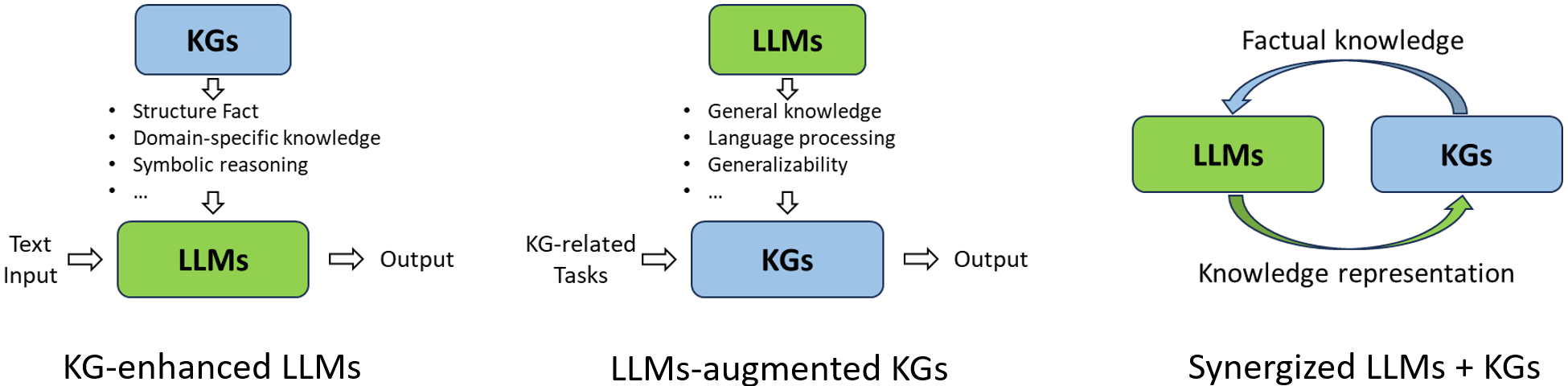}
	\caption{Three ways to combine LLMs with knowledge graph reasoning. Sourced from ~\cite{richardson2006markov}}
	\label{llm}
\end{figure*}

with the advent of ChatGPT, large language models have demonstrated great performance in many downstream tasks. Previously, we introduced knowledge graph reasoning. We know that knowledge graphs contain accurate structural knowledge and are very interpretable, however, most knowledge graphs are incomplete, lack language understanding. While language models are general knowledge, they are good at language understanding. However, they suffer from hallucination, lack interpretation, lacking new knowledge. By combining them together, we can build a model that is not only accurate but also interpretable. 

When combining knowledge graphs with large language models, there are three different ways. The first category is letting LLMs enhance knowledge graph reasoning, where LLMs serve as a component in the reasoning process. The second category is Knowledge graph reasoning enhance LLMs where knowledge graph reasoning can be used to mitigate the LLM’s hallucination problem. Finally, integrating knowledge graph reasoning with LLMs in a mutually beneficial way, so that they can help each other. Figure ~\ref{llm} shows the classification.

\subsection{Knowledge graph enhances LLMs.}  
Many methods have been proposed to utilize knowledge graph to boost the LLMs performance. In QA-GNN ~\cite{qagnn}, it combines LLM and knowledge graph to answer multi-choice questions. First of all, given a QA context, it will use the language model to encode question to a vector presentation, then use stand method to retrieve a knowledge graph subset, like linking entities and get their neighbors, and reasoning according to the subgraph. 
Then QA-GNN is based on two core ideas. First, in order to better identify which knowledge graph nodes relevant to current question, they propose language condition KG nodes scoring where they use a pretrained language model to compute the probability of each entity condition on the current question. 
Secondly, to jointly reason with language models, they connect the question text and kg to form a joint graph, which we call working graph, they mutually update their representations, through graph neural networks. Finally, they combine the representation of the language model and kg to predict the final answer. 
Following this direction, GreaseLM ~\cite{greaselm} merges LM with graph neural network by using a merging layer. To encode the knowledge graph structure information, multiple merging layers are used.

Now, we have introduced how to use knowledge graphs to help retrain language models to better answer different types of questions. However, when the size of the model becomes large, retrain or finetune the model will be very time consuming. An alternative way is to retrieval knowledge from external sources to help the language model generate correct answers. This approach allows for more targeted adjustments without the need for extensive retraining. This type of method is called Retrieval-Augmented Generation short for RAG. 

In KG-GPT ~\cite{kim2023kg}, the authors propose a method to utilize language models and knowledge graphs to answer more complex natural language questions. The idea is that a sentence of question, it uses llms to decompose the original sentence to several sub-sentences and find answers for each sub-sentence. And finally, find the answer for the whole sentence.
In REPLUG ~\cite{replug}, a new retrieval-augmented LM framework where the language model is viewed as a black box and the retrieval component is added as a tunable plug-and-play module. Given an input context, REPLUG first retrieves relevant documents from an external corpus. The retrieved documents are prepended to the input context and fed into the black-box LM to make the final prediction. Because the LM context length limits the number of documents that can be prepended, they also introduce a new ensemble scheme that encodes the retrieved documents in parallel with the same black-box LM. Then we pass the concatenation of each retrieved document with the input context through the LM in parallel, and ensemble the predicted probabilities. 
They have also developed REPLUG LSR. Instead of adjusting the language model to fit the retriever, they adapt the retriever to the language model. they train it to find documents that help improve the language model's understanding, while keeping the language model unchanged. The training process involves four steps: (1) finding and assessing the relevance of documents, (2) scoring these documents using the language model, (3) adjusting the retrieval model based on how different the retrieval and language model scores are, and (4) updating the index of the data in real-time.

Previous approaches of retrieval-augmented language models can only answer questions where answers are contained locally within regions of text and fail on answering global questions such as “what are the main themes in the dataset?” To solve these questions, the work proposes a method called GraphRAG ~\cite{edge2024local} which is a two-step process. The approach uses an LLM to build a graph-based text index in two stages: first to derive an entity knowledge graph from the source documents, then to pregenerate community summaries for all groups of closely-related entities. Given a question, each community summary is used to generate a partial response, before all partial responses are again summarized in a final response to the user.

\subsection{LLMs enhances knowledge graph reasoning.} 
Traditional methods retrieve information from KGs, augment the prompt accordingly, and feed the increased prompt into LLMs. In this paradigm, LLM plays the role of translator which transfers input questions to machine-understandable command for KG searching and reasoning, but it does not participate in the graph reasoning process directly. Its success depends heavily on the completeness and high quality of KG, which means that if the knowledge graph contains many missing edges, the answer won’t be good. So, recently, some people tried to explore how to mitigate the knowledge graph incompleteness problem by treating the LLM as an agent to travel KGs and perform reasoning based on paths. In Think-on-Graph ~\cite{sun2023think}, LLM is treated as an agent to traverse on knowledge graph to find answers. Now, let’s see an example, given question “What is the majority party now in the country where Canberra is located?” The llm identitied the topic entity is Canberra, by checking its local neighbourhood, Australia has the highest score. But the LLM notices that there is not enough information to get the answer. So it travels to Australia and check its neighbors. 

\subsection{LLMs and knowledge graph reasoning mutually help each other.}  
Finally, let us introduce the third part, how to let knowledge graph reasoning and large language models mutually help each other. 

One of the most important tasks in knowledge graph reasoning is to learn the embedding for the KG entities. It has been shown that the learned embedding from KG can be used to improve the performance of Pre-trained Language Models. On the other hand, for each node in the knowledge graph, we can find many text descriptions to describe it. This text information can be used to learn the node embedding. The strong ability of pre-trained language model can help us learn the embedding. So, in KEPLER ~\cite{kepler}, they propose a model which can also solve two problems at the same time. 
In this paper, Roberta is chosen as the encoder. Given a triplet, each node has a text description. Roberta is used to learn entity embedding. And the learned embedding will be used to calculate the knowledge graph embedding loss. On the other hand, conventional masked language model is used. It requires the model to predict the masked token within the text. The final loss is the summation of these two losses.

To leverage the structure information during the reasoning process, In ~\cite{jaket}, this work proposes to use a graph attention network to provide structure-aware entity embeddings for language modeling. More specifically, the language module produces contextual representations as initial embeddings for KG entities and relations given their descriptive text. Then, graph neural network is used to update the node embedding and relation embedding. 
Next, the learned knowledge graph embedding will be used as the input of the language model. And the output of the language model will be used to solve different downstream tasks.  There are several advantages of this method. For example, the joint pre-training effectively projects entities/relations and text into a shared semantic latent space, which eases the semantic matching between them.

%% file: 008conclusion.tex
\subsection{Conclusion}

Knowledge graphs serve as intuitive repositories of human knowledge, facilitating the inference of new information. However, traditional symbolic reasoning struggles with incomplete and noisy data often found in these graphs. Neural Symbolic AI, a fusion of deep learning and symbolic reasoning, offers a promising solution by combining the interpretability of symbolic methods with the robustness of neural approaches. Furthermore, the advent of large language models (LLMs) has opened new frontiers, enhancing knowledge graph reasoning capabilities.
We explore these advancements by categorizing knowledge graph reasoning into four main areas: single-hop queries, complex logical queries, natural language questions and LLMs with knowledge graph reasoning. For single-hop queries, we review techniques that efficiently handle direct relationships between entities. Complex logical query reasoning involves multi-hop inference, where we discuss methods that manage intricate relationships and multiple steps of reasoning. Finally, we delve into reasoning for natural language questions, including both single-turn and multi-turn dialogues. This survey aims to provide a comprehensive overview, bridging the gap between different reasoning approaches and offering insights into future research directions.

\subsection{Future Directions}

Despite significant progress in knowledge graph reasoning, several challenges remain unsolved. Current research primarily focuses on reasoning within a single knowledge graph, often overlooking the potential of integrating knowledge from different languages and modalities. To address these gaps, we outline future directions that could advance the field.

The first future direction is reasoning on multi-modal knowledge graphs. These graphs combine structured knowledge with unstructured data such as images, videos, and audio. Reasoning on multi-modal knowledge graphs can uncover valuable insights that single-modal data cannot. For example, it can predict whether two images contain the same building or associate text descriptions with relevant multimedia content. Integrating multiple data types enhances the robustness and applicability of knowledge graph reasoning, making it possible to address more complex and diverse queries.

The second future direction is reasoning on cross-lingual knowledge graphs. Many knowledge graphs exhibit similar patterns despite being described in different languages. Mining these patterns can reveal useful information that transcends linguistic boundaries. Potential research directions include language-agnostic representation learning, multilingual logical rule extraction, and cross-lingual link prediction. These approaches aim to create models that understand and reason across different languages, enabling more inclusive and comprehensive knowledge graph applications. This can significantly enhance global information retrieval, allowing for more effective use of knowledge graphs in multilingual contexts.